\RequirePackage[hyphens]{url} 
\documentclass[review]{elsarticle}
\usepackage{siunitx}
\usepackage{hyperref}
\usepackage{subfigure}
\usepackage{color, soul}
\usepackage{amsmath}
\usepackage{caption}
\usepackage{wrapfig}
\PassOptionsToPackage{hyphens}{url}

\journal{Neural Networks}

\newcommand{\ignore}[1]{}
\def\BibTeX{{\rm B\kern-.05em{\sc i\kern-.025em b}\kern-.08em
    T\kern-.1667em\lower.7ex\hbox{E}\kern-.125emX}}

\begin{document}

\begin{frontmatter}

\title{SpinAPS: A High-Performance Spintronic  Accelerator for Probabilistic Spiking Neural Networks}

\author[mymainaddress]{Anakha V Babu}

\author[mysecondaryaddress]{Osvaldo Simeone }

 
\author[mysecondaryaddress]{Bipin~Rajendran\corref{mycorrespondingauthor}}
\cortext[mycorrespondingauthor]{Corresponding author}
\ead{bipin.rajendran@kcl.ac.uk}

\address[mymainaddress]{Department of Electrical and Computer Engineering, New Jersey Institute of Technology, NJ, 07102 USA}
\address[mysecondaryaddress]{Department of Engineering, King’s College London, King’s College London, WC2R 2LS, UK}

\begin{abstract}
We discuss a high-performance and high-throughput hardware accelerator for probabilistic Spiking Neural Networks (SNNs) based on Generalized Linear Model (GLM) neurons, that uses binary STT-RAM devices as synapses and digital CMOS logic for neurons. 
The inference accelerator, termed ``SpinAPS'' for \textbf{Spin}tronic \textbf{A}ccelerator for \textbf{P}robabilistic \textbf{S}NNs, implements a principled direct learning rule for first-to-spike decoding without the need for conversion from pre-trained ANNs. The proposed solution is shown to achieve comparable performance with an equivalent ANN on handwritten digit and human activity recognition benchmarks. The inference engine, SpinAPS, is shown through software emulation tools to achieve $4\times$ performance improvement in terms of GSOPS/W/mm$^2$ when compared to an equivalent SRAM-based design. The architecture leverages probabilistic spiking neural networks that employ first-to-spike decoding rule to make inference decisions at low latencies, achieving $75\%$ of the test performance in as few as $4$ algorithmic time steps on the handwritten digit benchmark. The accelerator also exhibits competitive performance with other memristor-based DNN/SNN accelerators and state-of-the-art GPUs.
\end{abstract}

\end{frontmatter}

\section{Introduction}

The explosive growth of processing requirements of data-driven applications has resulted in intensive research efforts for alternative computing architectures that are more energy-efficient than traditional von Neumann processors. Unlike the dominant Deep Neural Networks (DNNs) which rely on real-valued information encoding, Spiking Neural Networks (SNNs) communicate through discrete and sparse tokens in time called spikes, mimicking the operation of the brain and are hence projected to be ideal candidates for realizing energy-efficient hardware platforms for artificial intelligence applications.  Moreover, SNNs are also ideal for real-time applications as they take advantage of the temporal dimension for data encoding and processing. However, SNNs lag behind
DNNs in terms of computational capability demonstrations
due to the current lack of efficient learning algorithms \cite{IEEE_nano}. Backpropagation techniques are ubiquitously adopted to train DNNs, but the discontinuous nature of spikes makes it non-trivial to derive such gradient-based rules for SNNs \cite{Roy2019}. 
Here we explore a probabilistic framework for SNNs, which defines the output of spiking neurons as jointly distributed binary random processes. Such definitions help in applying maximum likelihood criteria and then derive flexible learning rules without requiring backpropagation mechanisms, conversions or other approximations from pre-trained ANNs \cite{jang_IEEE_mag}. 

Conventional neural networks have millions of trainable parameters and are
trained using von Neumann machines, where the memory and computation units are
physically separated. The performance of these implementations is typically limited
by the ``von Neumann bottleneck" caused by the constant transfer of data between
the processor and the memory. SNN implementations on these platforms becomes
inherently slow due to the need to access data over time in order to carry out temporal
processing. Hence, hardware accelerators are necessary to exploit the full potential of
SNNs and also to develop efficient learning algorithms. Here, we discuss a hardware accelerator designed for implementing probabilistic SNNs, that uses binary STT-RAM devices for realizing synapses and digital CMOS for neuronal computations. We also evaluate the performance of probabilistic SNNs against equivalent ANNs on standard benchmark datasets. 


\section{Related Work}
\label{Sec:RelatedWork}
In an effort to build large-scale
neuromorphic computing systems that can emulate the energy efficiency of the human brain, several computing platforms have implemented SNNs. While Tijanic chip \cite{Tijanic}, Intel's Loihi \cite{intel_loihi} and IBM's TrueNorth \cite{TrueNorth} realize spiking neurons and make use of static random access memory (SRAM) to store the state of synapses and neurons, recent research efforts have  proposed the use of tiled crossbar arrays of two terminal nanoscale devices for implementing large-scale DNN systems \cite{Kuzum_2013, RPU, Ambrogio2018, PUMA}. Though analog memristor based neural network accelerators are estimated to provide higher throughput than GPUs, there are several challenges associated with these in-memory computing solutions \cite{RPU, PUMA, ISSAC, PRIME}. For instance, programming variability and stochasticity of device conductances as well as the need for accurate and area/power-hungry digital to analog converters (DAC) at the input and analog to digital converters (ADC) at the output, along with the additive noise contributed by these peripheral logic circuits pose significant  challenges \cite{Ambrogio2018, Anakha_neruo_comp, SRK_STT_RAM}. 

Among the emerging nanoscale memory devices, Spin Transfer Torque RAM (STT-RAM) has  been explored for implementing synaptic weights for ANNs as well as SNNs owing to their fast read/write characteristics, high endurance, and scalability \cite{STTRAM_synapse, STT_RAM_synapse1} in several previous studies  \cite{SRK_STT_RAM, Sengupta_IJCNN, SRK_STT_RAM_learn, STT_RAM_ANNs}. The stochastic nature of the spintronic devices has been leveraged to model neural transfer functions in a crossbar architecture  \cite{STTRAM_prob_SNN}. These designs for SNNs have shown over $20\times$ energy improvements over conventional CMOS designs. 
Notably, an all-spin neuromorphic processor design for SNNs comprising of spintronic synapses and neurons with in-memory computing architecture has been projected to give $1000\times$ higher energy efficiency and $200\times$ speed up compared to  equivalent CMOS implementations  \cite{Sengupta_IJCNN}. 
In contrast to these prior implementations using spintronic devices, we consider STT-RAM device as a binary storage unit, minimising the area/power and precision requirements of peripheral logic circuits consisting of ADCs/DACs. Thus, $n$-bit synaptic weight is  represented by using $n$ binary STT-RAM devices. While the idea of using STT-RAM device as a binary storage unit has been discussed earlier in \cite{SRK_STT_RAM, SRK_STT_RAM_learn}, a deterministic spiking neuron model was used for learning and inference based on a crossbar architecture. Instead, this work uses STT-RAM memory as a regular storage unit, similar to that of SRAM storage in conventional designs. To the best of our knowledge, this work  presents the  first design of a hardware accelerator based on STT-RAM devices to implement probabilistic SNNs based on a generalized linear model (GLM) for neurons. 


\subsection{Main Contributions}
The main contributions of this paper are listed as follows. 
\begin{itemize}
    \item We propose a hardware accelerator for inference, \textbf{Spin}tronic \textbf{A}ccelerator for \textbf{P}robabilistic \textbf{S}NNs (SpinAPS) that integrates binary spintronic devices to store the synaptic states and digital CMOS neurons for computations. The potential of hardware implementations of Generalized Linear Model (GLM)-based probabilistic SNNs trained using the energy efficient first-to-spike rule are evaluated for the first time in this work.
    \item We evaluate the performance of  probabilistic SNNs on two benchmarks - handwritten digit recognition and human activity recognition datasets, and show that  SNN performance is comparable to that of  equivalent ANNs. 
    \item SpinAPS achieves $4\times$ performance improvement in terms of GSOPS/W/mm$^2$   when compared to an equivalent design that uses  SRAM  to store the synaptic states.
\end{itemize}

This paper is organized as follows: In Section \ref{sec:GLM_arch}, we review the architecture of GLM-based probabilistic SNNs and also explain the first-to-spike rule for classification. The algorithm optimization strategies for an efficient hardware implementation for the first-to-spike rule is considered in Section \ref{sec:alg_opt}. The architecture details of SpinAPS core and the mapping of the GLM kernels into the memory is discussed in Section \ref{sec:core}. Section \ref{sec:full_design} details the relevant digital CMOS neuron logic blocks and memory design. We then evaluate the performance of our hardware accelerator for different bit precision choices in Section \ref{sec:perf_eval}. Finally, we conclude the paper in Section \ref{sec:conclude}.  

\section{Review of GLM neuron model for SNNs}
\label{sec:GLM_arch}
The majority of the neuron models used in SNN computations are deterministic in nature, by which the neuron emits a spike when the membrane potential crosses the threshold value. The non-differentiable nature of the spikes in SNNs makes it non-trivial to use the standard approach of stochastic gradient descent (SGD) widely used for training ANNs. Therefore, several approaches have been used to realize deterministic SNNs such as converting pretrained ANNs, smoothing out the membrane potential to define the derivatives, and so on \cite{SNN_realize1, SNN_realize2, SNN_realize3, SNN_realize4}. In contrast to deterministic models, probabilistic models for neurons are based on a linear-nonlinear Poisson model that are widely studied in the field of computational neuroscience \cite{Pillow_GLM}. 

\begin{figure}[!ht]
	\centering
 	\includegraphics[width=0.9\textwidth]{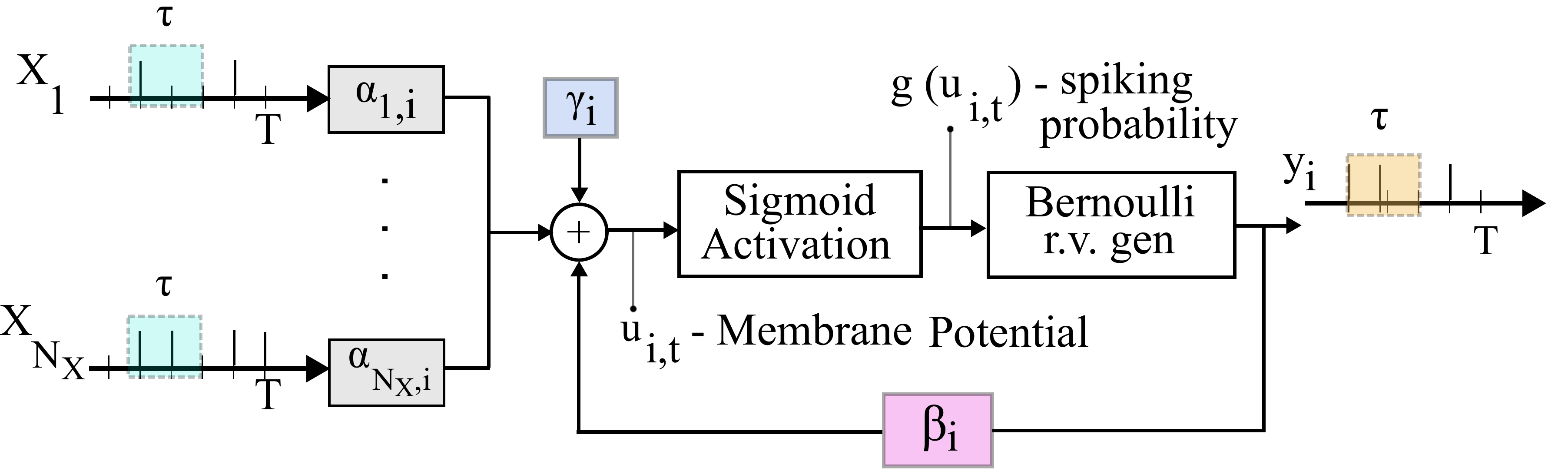}
	\caption{The architecture of Generalized Linear Model (GLM) model used for SNN learning in this work \cite{Alireza_FTS}. $N_X$ input neurons and one ouput neuron is shown for simplicity.}	
	\label{fig:GLM_architecture}
\end{figure}

Generalized Linear Models (GLM) yield a probabilistic framework for SNNs that is flexible and computationally tractable \cite{jang_IEEE_mag, Pillow_GLM}. The basic architecture of GLM-based probabilistic SNNs used for SNN training is shown in Fig. \ref{fig:GLM_architecture}. We focus on a  2-layer SNN, which has $N_X$ presynaptic neurons encoding the input and $N_Y$ output neurons corresponding to the output classes. Each of the input neurons receives a spike train having $T$ samples through rate encoding. The input is normalized and spikes are issued through a Bernoulli random process. For cases where the
sign of the input is vital in achieving learning, the negative sign is absorbed in the sign bit of the corresponding weights. The membrane potential of $i^{th}$ output neuron at any instant $t$ can be expressed as, 
\begin{equation}
\label{eq:mem_pot}
		u_{i,t}= \sum_{j=1}^{N_X}\alpha_{j,i}^Tx_{j,t-\tau_y}^{t-1} + \beta_{i}^Ty_{i,t-\tau_y}^{t-1} +\gamma_i 
		\end{equation}
where $\alpha_{j,i}$ denotes the stimulus kernel; $x_{j,t-\tau}^{t-1}$ represents the input spike window having $\tau$ spike samples, $\beta_{i}$ denotes the feedback kernel; ${y_{i,t-\tau}^{t-1}}$ represents the output spike window with $\tau$ samples; and $\gamma_i$ is the bias parameter. Here $j$ refers to the index of the pre-synaptic neuron and $i$ refers to the index of the post-synaptic neuron.  The stimulus and feedback kernels in GLM can be defined as the weighted sum of fixed basis functions with the learnable weights, and they are expressed as shown below. 
\begin{equation}
		\alpha_{i,j}  = \textbf{A}\textbf{w}_{i,j}\text{, } \hspace{2 mm}
		\beta{i}  = \textbf{B}\textbf{v}_{i}
	\end{equation}
The matrices $\textbf{A}$, $\textbf{B}$ are the basis vectors,  defined as $\textbf{A}$ = [$a_1,\ldots, a_{K_{\alpha}}$] and $\textbf{B}$ = [$b_1, \ldots, b_{K_{\beta}}$]. The prior work in GLM \cite{Alireza_FTS} uses real-valued raised cosine basis vectors for the stimulus and the feedback kernels. The vectors $\textbf{w}_{j,i}$ = $[w_{j,i,1},\ldots , w_{j,i,K_{\alpha}}]^T$ and $v_{i} = [v_{i,1},\ldots , v_{i,K_{\beta}}]^T$ are the learnable weights in the network with $K_{\alpha}$, $K_{\beta}$ denoting the number of basis functions. 
The spiking probability of the output neuron is then decided based on the sigmoid non-linearity applied to the membrane potential. 
GLM neurons have reproduced a wide range of spiking neuronal behaviors observed in human brain by appropriately tuning the stimulus and feedback kernels \cite{Pillow_GLM}.
Learning rules for a GLM SNN based on rate and first-to-spike decoding rules have been derived in a number of works reviewed in \cite{ jang_IEEE_mag,Jang_Osvaldo, gardner2016supervised}.

Two main decoding strategies have been considered for the given SNN architecture - rate decoding and first-to-spike decoding \cite{Alireza_FTS}. When using the rate decoding scheme for inference, the network decision is based on the neuron with the maximum spike count. For the first-to-spike scheme, a decision is made when one of the output neuron spikes. It has been shown in \cite{Alireza_FTS} that the first-to-spike rule exhibits a low inference complexity compared to rate decoding due to its ability to make decisions early. Hence, we choose the first-to-spike scheme for our hardware optimization studies.

\subsection{First-to-Spike Decoding}
\begin{figure}[!ht]
	\centering
 	\includegraphics[width=0.9\textwidth]{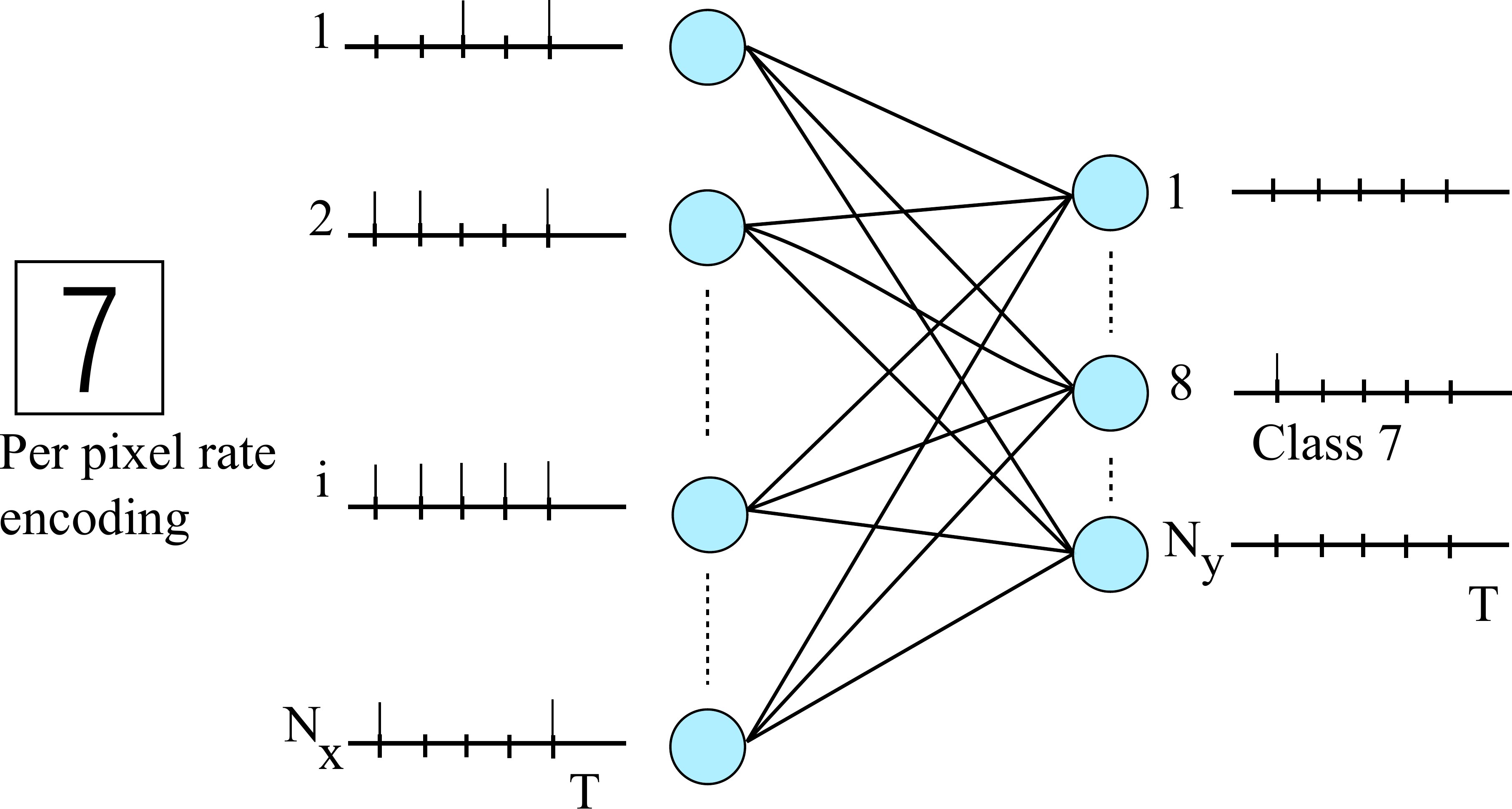}
		\caption{Illustration of first-to-spike (FtS) rule decoding scheme for the GLM-based SNN on the handwritten digit benchmark. For example, when the trained network is presented with an image `7', output neuron corresponding to class 7 will ideally generate the first spike. }
	\label{fig:GLM_f2s}
\end{figure}

The fundamental idea of first-to-spike scheme is illustrated in Figure \ref{fig:GLM_f2s}. The kernels for the GLM based SNN are trained using the maximum likelihood criterion, which maximizes the probability of obtaining the first spike at the labeled neuron and no spikes for all other output neurons up to that time instant. This probability can be mathematically expressed as, 
\begin{equation}
	p_t(\theta)= \prod_{i=1, i\neq c}^{N_Y} \prod_{t'=1}^{t} \overline{g}(u_{i,t'}) g(u_c,t) \prod_{t'=1}^{t-1} \overline{g}(u_{c,t'})
	\label{eq: FTS_prob}
	\end{equation} 
where $c$ corresponds to the labeled neuron, $u$ denotes the membrane potential, and $g(u)$ denotes the sigmoid activation function applied to the membrane potential $u$. Also  $\overline{g}(u)=1-g(u)$.	The weight update rules are derived by maximizing the log probability in equation (\ref{eq: FTS_prob}) and are discussed in detail in \cite{Alireza_FTS}. Note that the feedback kernels are not necessary for the first-to-spike rule as the network dynamics need not be calculated after the first spike is observed.
	
\subsection{Datasets Used in This Study}
\label{subsec:datasets}
Throughout this work, we evaluate the performance of Probabilistic SNNs on handwritten digit recognition and human activity recognition (HAR) datasets \cite{HAR}. With $60,000$ training and $10,000$ test images, each of the input image in the handwritten digit database has $784$ ($28\times28$) pixels corresponding to the $10$ ($0, 1, \ldots, 9$) digits. The HAR dataset has a collection of physical activities feature extracted from the time series inputs of the embedded sensors in a smart phone. The database has roughly $7,000$ training samples and $3,000$ test samples corresponding to   six types of physical activities.
\section{Hardware-Software Co-optimization}
\label{sec:alg_opt}
\begin{figure}[h!]
\centerline{\includegraphics[scale=0.35]{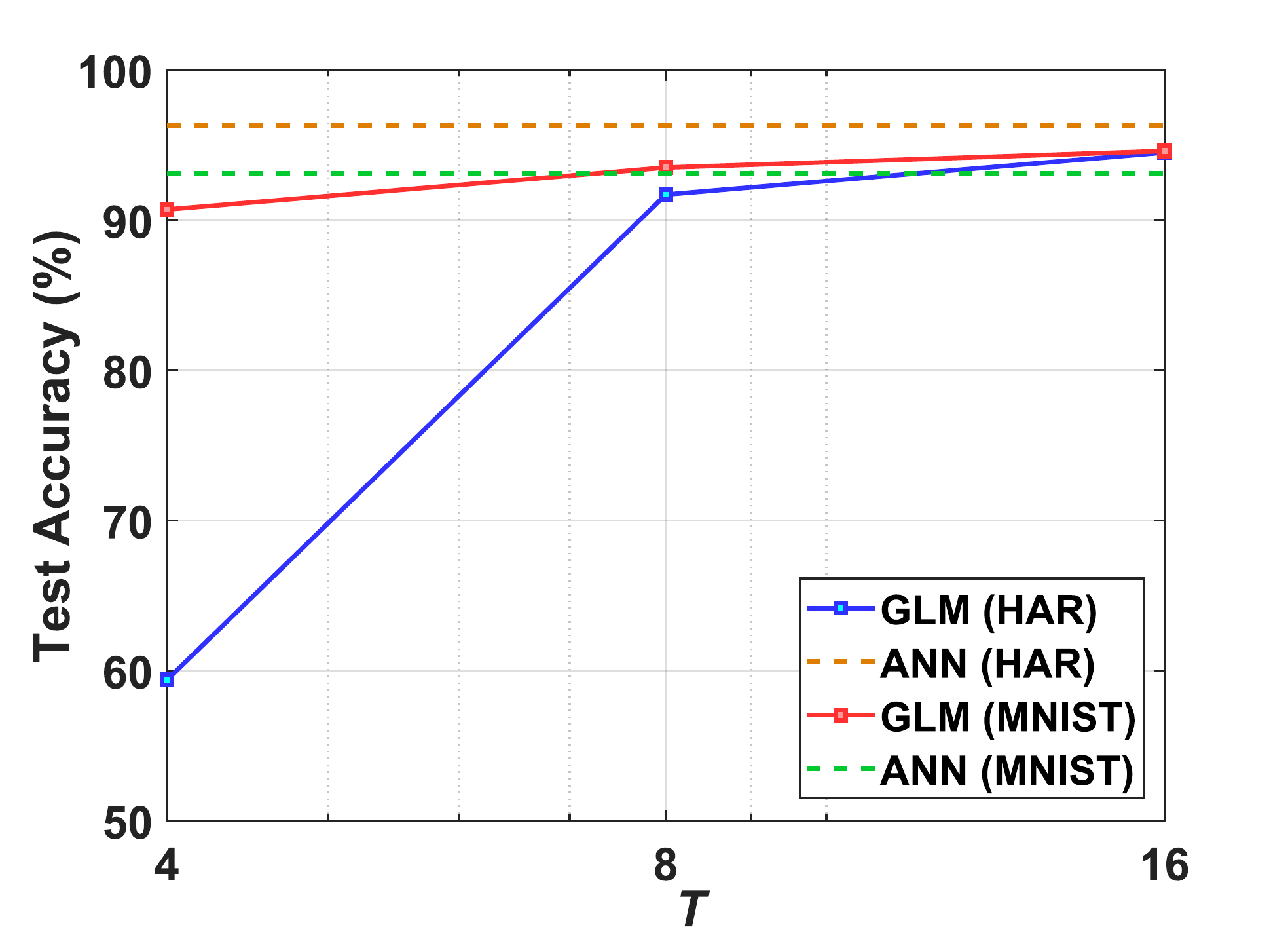}}
\caption{The test accuracy of GLM on human activity recognition (HAR) and handwritten digit recognition. 
A comparable performance is achieved with respect to an ANN having the same architecture. 
Here presentation time $T$ is kept the same as the spike integration window $\tau$.}
\label{fig:Test_acc}
\end{figure}

In this section, we discuss the design choices that facilitate our hardware implementation. We start by observing the floating-point baseline accuracy of FtS in Fig. \ref{fig:Test_acc}, where the kernels are trained using the techniques  demonstrated in \cite{Alireza_FTS}. For the purpose of designing the inference engine, we assume that the kernels are fixed and binary basis vectors are used in the training instead of the cosine basis vectors. The usage of binary basis vectors helps in simplifying the kernel computations without requiring any multipliers in hardware. We first determine the floating baseline test accuracy as a function of the presentation times $T$ by training the SNN over $200$ epochs.

It can be seen that the network performance improves with higher presentation times $T$ and spike integration windows $\tau$, reaching a maximum test accuracy for $T=16$ and $\tau=16$. 
Note that for the handwritten digit benchmark, the network accuracy of the GLM SNN trained using the FtS rule with ($T=8$, $\tau=8$) is $93.5$\%, which is at par with the $93.1$\% accuracy of a 2-layer artificial neural network (ANN) with the same architecture. In the case of HAR dataset, GLM achieves a maximum test accuracy of $94.5\%$ with $T=16$, $\tau=16$ which is comparable with ANN test accuracy of $96.3\%$. Hence, we chose  $T=8$, $\tau=8$ for handwritten digit recognition and $T=16$, $\tau=16$ for  HAR benchmarks as the baseline. We now discuss software optimization strategies for implementing the first-to-spike scheme in an energy efficient manner in hardware.

\subsection{Sigmoid Activation Function}
\label{subsec:PWL1}
As shown in Fig. \ref{fig:GLM_architecture}, the basic GLM neuron architecture uses a sigmoid activation function to determine the spike probability of the output neurons. The implementation of sigmoid functions in hardware is relatively complex as it involves division and exponentiation, incurring significant  area and  power \cite{Sigmoid_bkgnd}. Here we follow the piecewise linear approximation (PWL) demonstrated in \cite{Sigmoid_PWL} for implementing the sigmoid activation function. In the PWL approximation, the sigmoid is broken up into integer set of break points such that the resulting function can be expressed as powers of $2$. Considering the negative axis alone, the PWL approximation $y$ for input ${x}$ can be expressed as 
\begin{equation}
\label{eq:PWL}
y_{x<0}=\frac{\frac{1}{2}+\frac{\hat x}{4}}{2^{\mid {(x)\mid}}}
\end{equation}
 where ${(x)}$ is the integer part of $x$ and $\hat{x}$ is the fractional part of $x$. 
 Using the symmetry of sigmoid function, the values in the positive x-axis can be obtained as $y_{x>0} =1-y_{x<0}$. 
The output of the PWL approximation is then compared with a pseudo-random number generated from the linear feedback shift register (LFSR) and a spike is generated if the PWL value is greater than the LFSR output. A 16-bit LFSR is used and is assumed to be shared among all the output neurons.

\subsection{Quantization}
\begin{figure}[h!]
\centerline{\includegraphics[width=\linewidth]{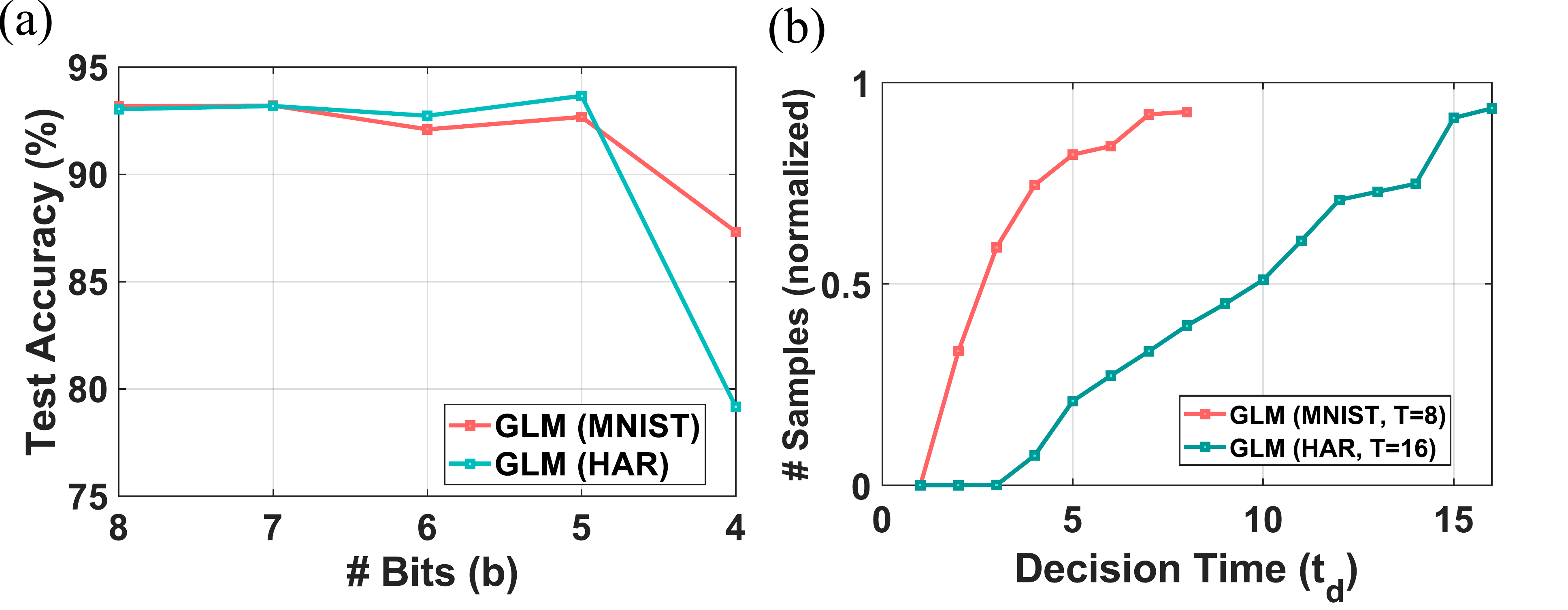}}
\caption{(a) Test performance of the GLM SNN after quantization of weights with PWL approximation and by using a 16-bit LFSR. (b) Cumulative distribution of the number of input samples classified as a function of decision time $t_d$. An early decision can be made in first $4$ time steps for classifying 75\% of the images in the handwritten digit benchmark.}
\label{fig:Qtzn_lat}
\end{figure}

We now aim at finding the minimum bit precision required for the learnt weights $w_{j,i}$ and biases $\gamma_i$ in order to maintain close to baseline floating-point accuracy during inference. Here we follow a post training quantization study to determine the minimum bit precision required for the learnable parameters.   
Starting with the baseline networks for the two datasets  obtained using floating point parameters, we quantize them into discrete levels and study the inference accuracy  with the PWL approximation for the sigmoid. With b-bit quantization, one bit is reserved for the sign to represent both positive and negative parameters. We use a uniform quantizer with quantization step given by the relation
\begin{align}
\label{eq:quant}
    w_{q}= \frac{w_{max}-w_{min}}{2^{\text{b}-1}},  \text{  }
    \gamma_{q} = \frac{\gamma_{max}-\gamma_{min}}{2^{\text{b}-1}} \text{, }
\end{align}
respectively for weights and biases.
We also quantize the membrane potential and the output of the PWL activation to b bits. We summarize the inference performance of the GLM SNN as a function of bit precision in Fig. \ref{fig:Qtzn_lat} (a).
Network performance degrades with lower choices of b, but we note that $5$ bit resolution is sufficient for maintaining  close to baseline floating-point inference accuracy for the two benchmarks.
Even though the input spike pattern lasts for $T$ algorithmic time steps, the first-to-spike learning rule allows a decision to be made even before all the spikes are presented to the network.
It can be observed for the handwritten digit benchmark, around 75\% of the samples are classified in the first $4$ algorithmic time steps. Thus GLM SNN can leverage the ability of first-to-spike rule in making decisions with reduced latency and memory accesses.
\section{Overview of SpinAPS architecture}
\label{sec:core}
\begin{figure}[h!]
\centerline{\includegraphics[scale=0.35]{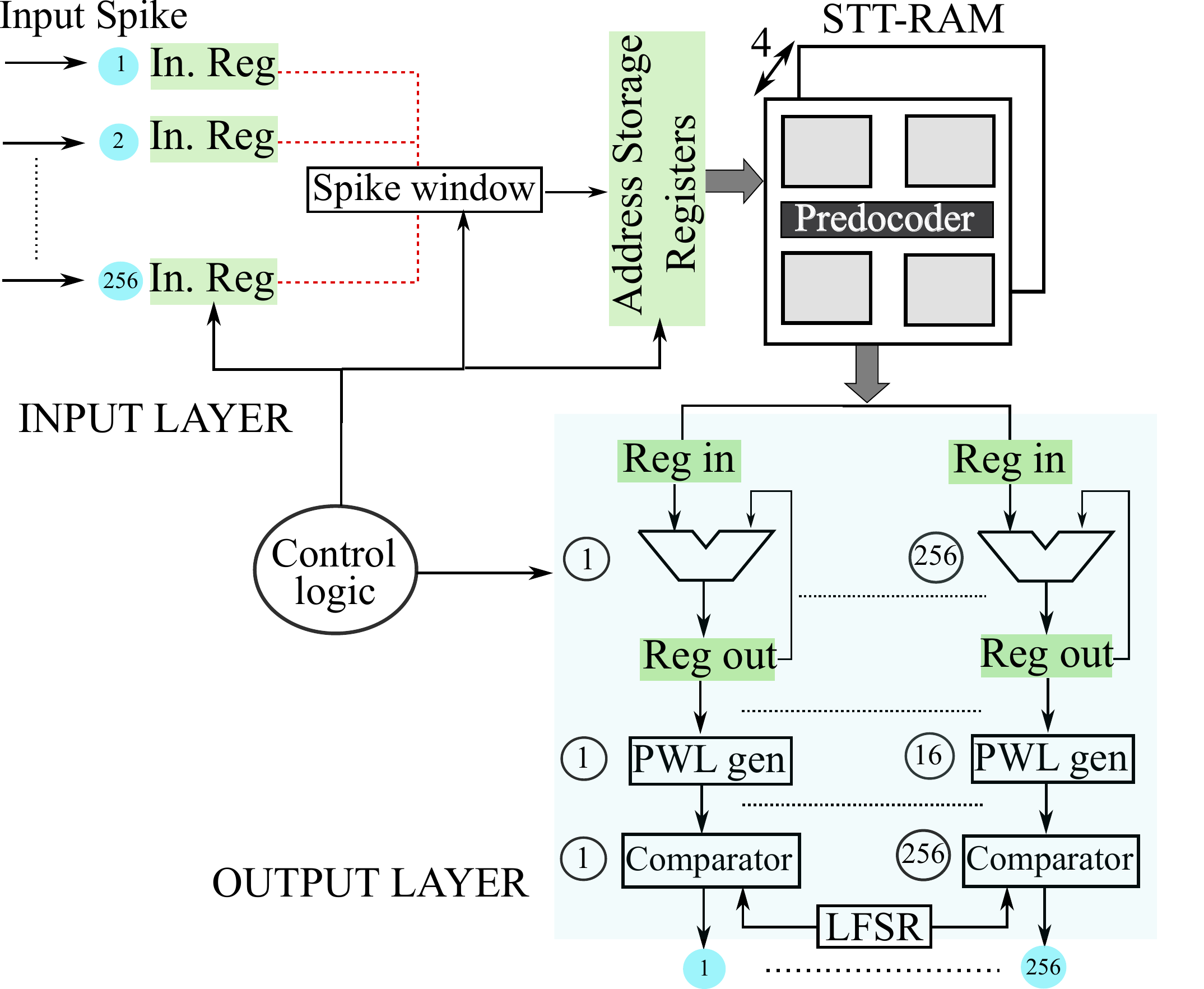}}
\caption{The core architecture of SpinAPS having binary STT-RAM devices to store the synaptic states and digital CMOS neurons. Assuming a spike integration window of $\tau=7$, the core can map $256$ input and $256$ output neurons.}
\label{fig:core}
\end{figure}

\begin{figure}[h!]
\centerline{\includegraphics[scale =0.35]{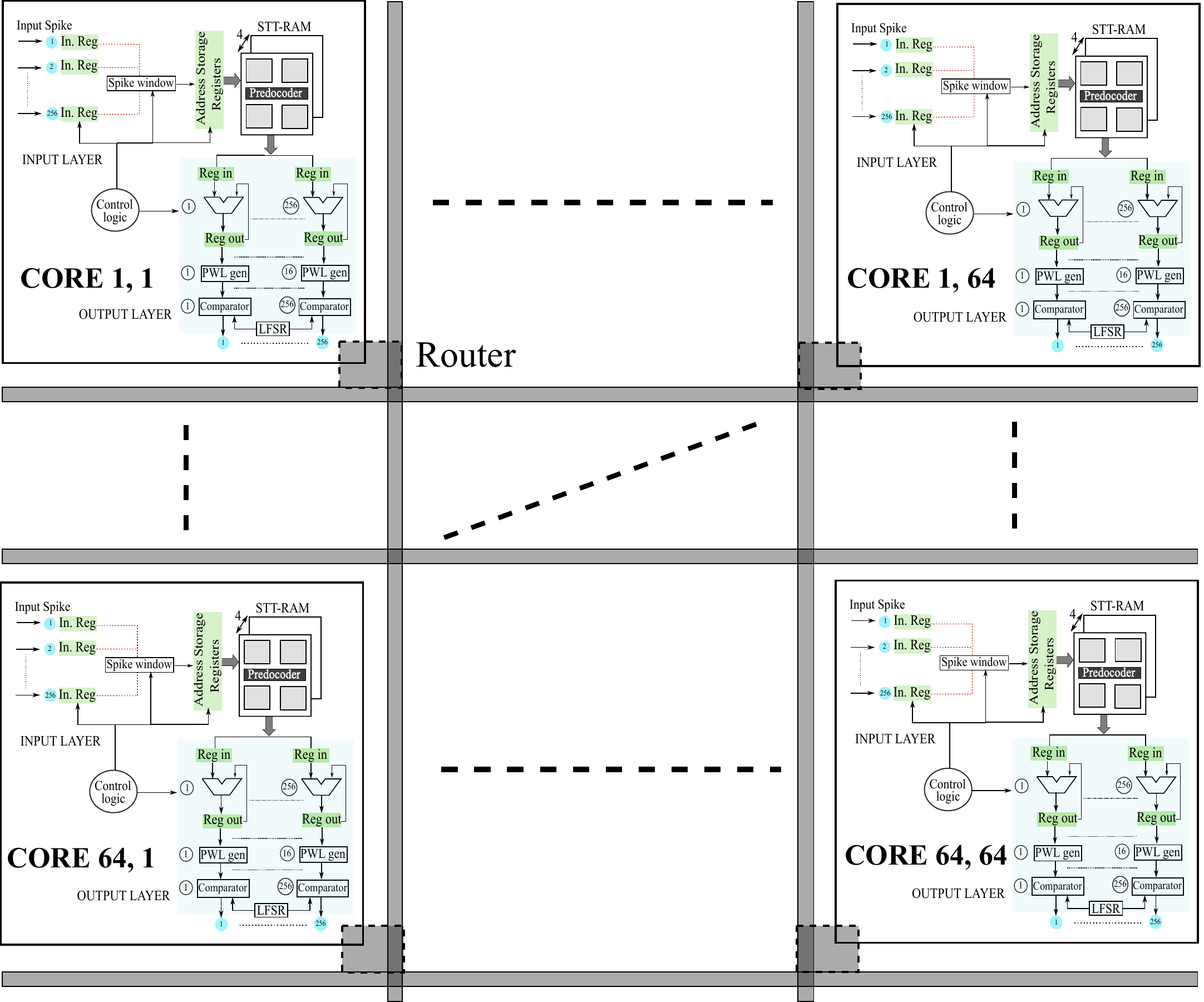}}
\caption{The tiled architecture of SpinAPS obtained by arranging the neuro-synaptic cores that communicate to each other using a packet routing digital mesh network.}
\label{fig:tile_arch}
\end{figure}

As illustrated in Fig. \ref{fig:core}, the core architecture of SpinAPS consists of binary spintronic devices to store the synaptic states and digital CMOS neurons to perform the neuronal functionality. The SpinAPS core accepts input spike at every processor time step $t$, reads the synapses corresponding to the spike integration window $\tau$ from the memory to compute the membrane potential and applies the non-linear PWL activation function to determine the spike probability of the output neurons. A pseudo-random number generated from the LFSR is then used to actually determine whether a spike is issued or not as explained in Section \ref{sec:alg_opt}. The word width, and hence the memory capacity of the banked STT-RAM memory used in the SpinAPS core can be designed based on the bit precision b required for the synapses.
SpinAPS co-locates the dense STT-RAM memory array and digital CMOS computations in the same core, thereby avoiding the traditional von Neumann bottleneck. Following the principles demonstrated in IBM's TrueNorth chip \cite{TrueNorth}, a tiled architecture with  $4096$ SpinAPS cores, with each core having $256$ neurons, can be used to realize a system with 1 million neurons  as illustrated in Fig. \ref{fig:tile_arch}.
We next discuss the basic characteristics of the spintronic synapses and how we can map the parameters of probabilistic SNNs into the memory.

\subsection{STT-RAM as Synapse}
 \begin{figure}[htbp!]
\centerline{\includegraphics[width=\linewidth]{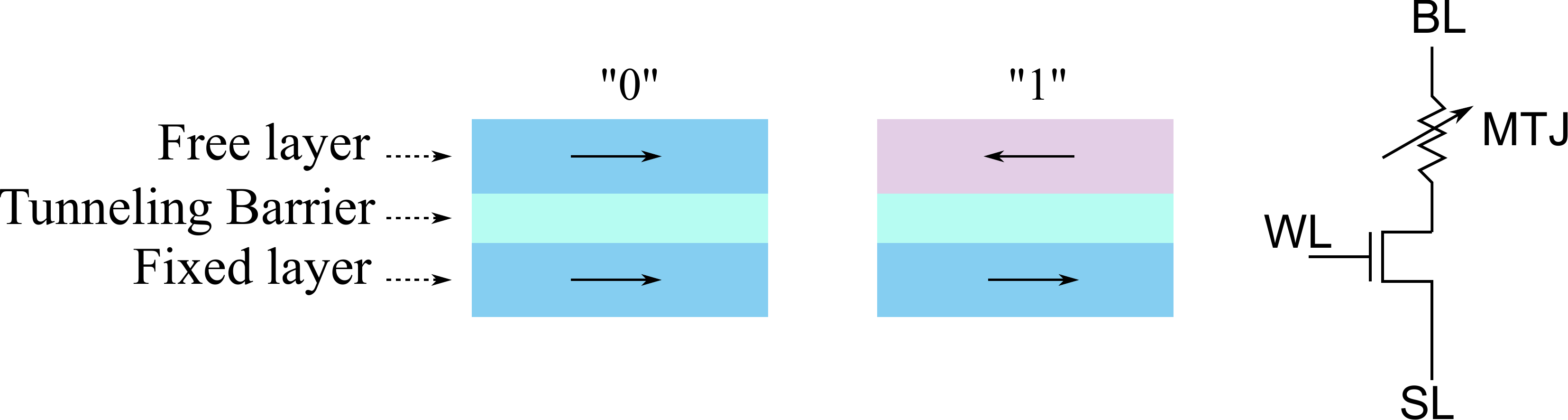}}
\caption{Illustration of basic cell structure of 1T/1MTJ STT-RAM with low (``0'') and high resistance (``1'') states. }
\label{fig:STT-RAM}
\end{figure}
 Synapses generally scale as $N^2$, where $N$ is the number of neurons in each layer. Recently, spintronic devices have been explored for its use as nanoscale synapses \cite{STTRAM_synapse, STTRAM_prob_SNN, STT_RAM_synapse1} mainly due to its high read and write bandwidths, low power consumption and excellent reliability \cite{STT-RAM_design, STT_RAM_40nm, STT_RAM_90nm}. These spintronic devices are compatible with CMOS technology and fast read/write has been demonstrated in sub $10$ ns regime \cite{STT_RAM_fastread, STT_RAM_3_3ns}. 
  Here we propose binary STT-RAM devices as the nanoscale electrical synapses for the SpinAPS core illustrated in Fig.  \ref{fig:core}. Structurally, STT-RAM uses a Magnetic Tunnel Junction (MTJ) with a pair of ferromagnets separated by a thin insulating layer. These devices have the ability to store either `0' or `1' depending on the relative magnetic orientation of the two ferromagnetic layers as shown in Fig. \ref{fig:STT-RAM}. The memory array is typically configured based on the crossbar architecture with an access transistor connected in series with the memory device to selectively read and program it.
  
  \subsection{Synaptic Memory Architecture}
\label{subsec:mem_bank}
Here we discuss the mapping of the learnable parameters, stimulus kernel ($\alpha$) and the bias parameters ($\gamma$)  of GLM-based SNN into the spintronic memory to achieve accelerated hardware performance.
\subsubsection{Kernel Mapping}
\begin{figure}[htbp!]
\centerline{\includegraphics[width=\linewidth]{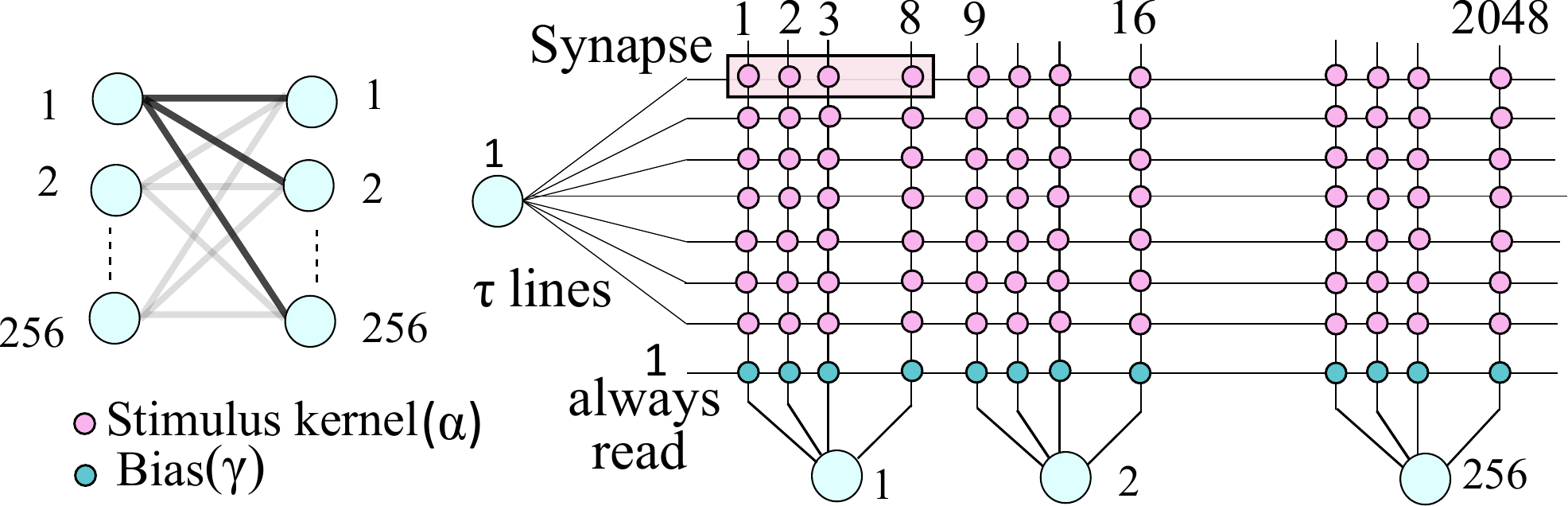}}
\caption{Kernel mapping of GLM SNN to the spintronic memory of SpinAPS core, showing that devices on $\tau$ word-lines are used to represent the kernel parameters of every input neuron.}
\label{fig:mem_mapping}
\end{figure}
Here we choose ($T=8$, $\tau=7$) to describe the mapping of $8$-bit synapses, hence the word width (vertical lines in Fig. \ref{fig:mem_mapping}) required for the memory is $256\times8=2048$ bits. Each of the input neuron generates a bit pattern of length $\tau$, hence unique kernel weights are provided for every neuron, requiring  $256\times 7=1792$  word lines in the array. Thus, a network with 256 input and 256 output neurons can be mapped to a memory array with $2048\times 2048$ memory devices. For example if the bit pattern generated by the first input neuron is ``1010010'', then the synaptic weights corresponding to 1$^{st}$, 3$^{rd}$, and 6$^{th}$ word line will be read sequentially. The address corresponding to these word lines are stored in registers, indicated as address storage registers in Figure \ref{fig:core}. These synaptic weights will be added at the  output neuron to compute the membrane potential. One word line will be sufficient for mapping $\gamma$ as ($256\times8$) bits is equivalent to the word width of the core. As $\gamma$ determines the baseline firing rate of the output neurons, the word line voltage corresponding to the $\gamma$ line is kept high indicating an ``always read'' condition. 

\section{High-level Design}
\label{sec:full_design}
\subsection{Neuron Implementation}
\label{sec:neuron_impl}
We now describe the  digital CMOS implementation of the relevant blocks. The baseline design assumes 8-bit precision and is synthesized using TSMC $45\,$nm logic process running at a clock frequency of $500\,$MHz.
\subsubsection{Generation of spike window}
\ignore{\begin{figure}[h!]
\centerline{\includegraphics[width=\linewidth]{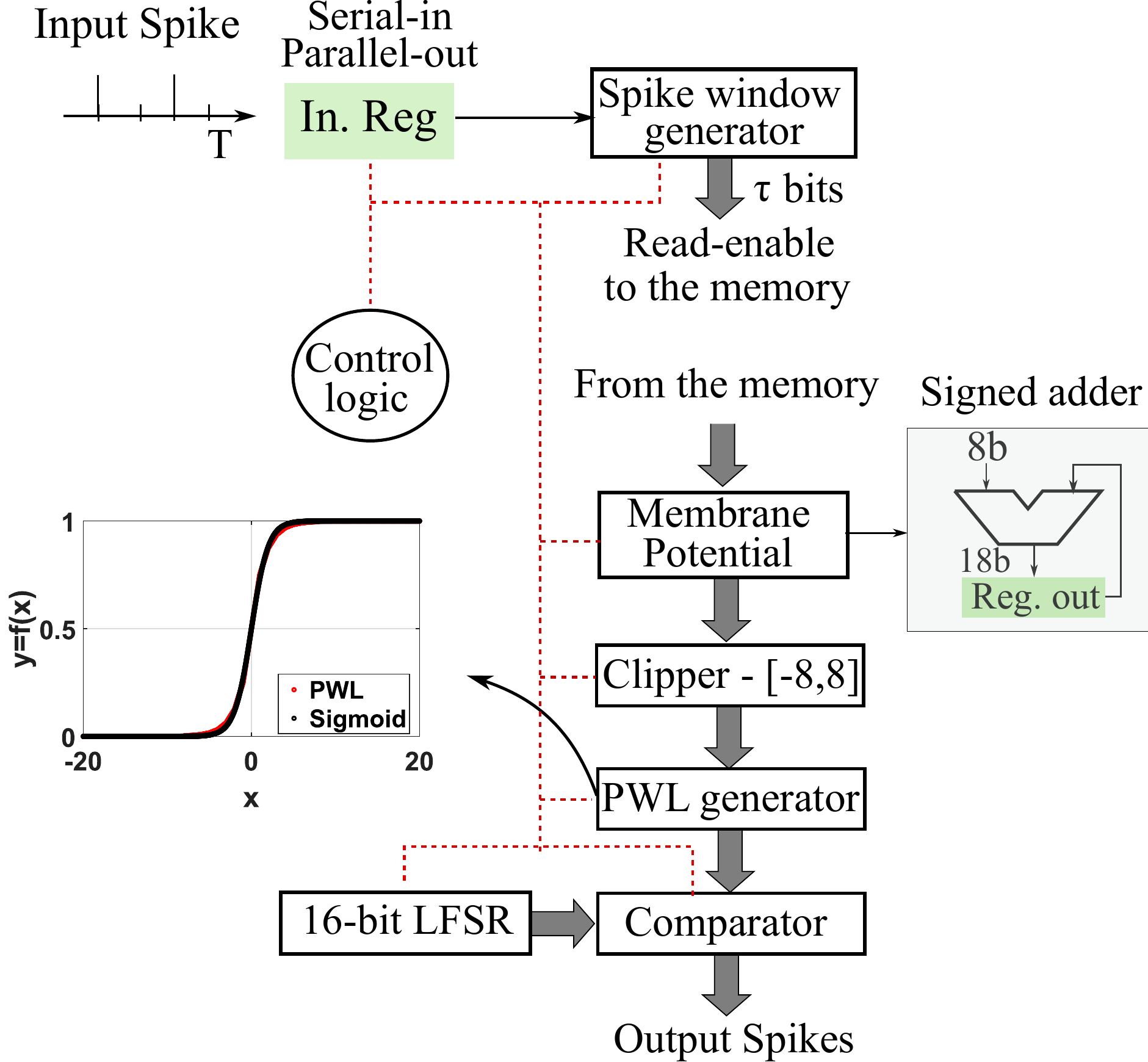}}
\caption{The digital CMOS logic blocks of SpinAPS core which is used to perform the probabilistic computations. Here we show the computations associated with one of the neurons in the core.}
\label{fig:core1}
\end{figure}}
As discussed in Section \ref{sec:GLM_arch}, normalized input spikes are transformed to generate spikes of length $T$ using a Bernoulli random process.
At each time instant, $t=1 \ldots T$, incoming spikes are latched at the input neuron (marked as `In. Reg', a serial-in, parallel out shift register in Fig. \ref{fig:core}) which is then translated into an activation pattern of length $\tau$, and applied as  read enable signals to the word lines associated with each kernel weights. The spike window generator circuit uses a multiplexer for selecting the bit pattern of length $\tau$ from the input register and whose select lines are configured by the controller for every $t$. For example, when $t=1$, the select line will be ``000'' and as no input spikes have arrived before, the output of the multiplexer would be ``0000000''.  When $t=3$, the select line becomes ``010'', and the multiplexer  output will be ``$s_2s_1$000000'', where $s_i=1/0$ represents the presence/absence of a spike at $t=i$ and so on. As discussed in Section \ref{sec:core}, the synapses associated with the input neurons are read sequentially and hence the logic circuit for the spike window generator can be shared among all the input neurons. The synaptic weights are stored in the registers and the sign bit of the weights can be optionally flipped depending on the sign of the input before computing the membrane potential $(u)$.

\subsubsection{Piecewise Linear Approximation (PWL) for sigmoid calculation}
\label{subsec:PWL}
Our GLM neuron model uses a non-linear sigmoid activation function for generating the spike probabilities for the output neuron based on the value of $u$.  
Here we adopt the traditional piecewise linear  approximation (shown in Equation \ref{eq:PWL}) for the sigmoid using adders and shift registers as described in \cite{Sigmoid_PWL}.
The 18-bit membrane potential $u$ is clipped to an 8-bit fixed point number in the range [-8,8] before applying to the Piecewise Linear Approximation (PWL) sigmoid generator which is shared among $16$ output neurons.  The clipped fixed point representation assumes $1$ bit for the sign, $4$ bits for the integer and $3$ bits for the fractional part. The output of the sigmoid generator is an unsigned $8$-bit number whose values lie in the range $[0,255]$ and is compared with the 8-bit pattern from the $16$-bit LFSR (Linear Feedback Shift Register) to generate output spikes.
\subsection{Design of the synaptic memory}
\label{subsec:design_syn_mem}
\begin{figure}[htbp!]
\centerline{\includegraphics[width=\linewidth]{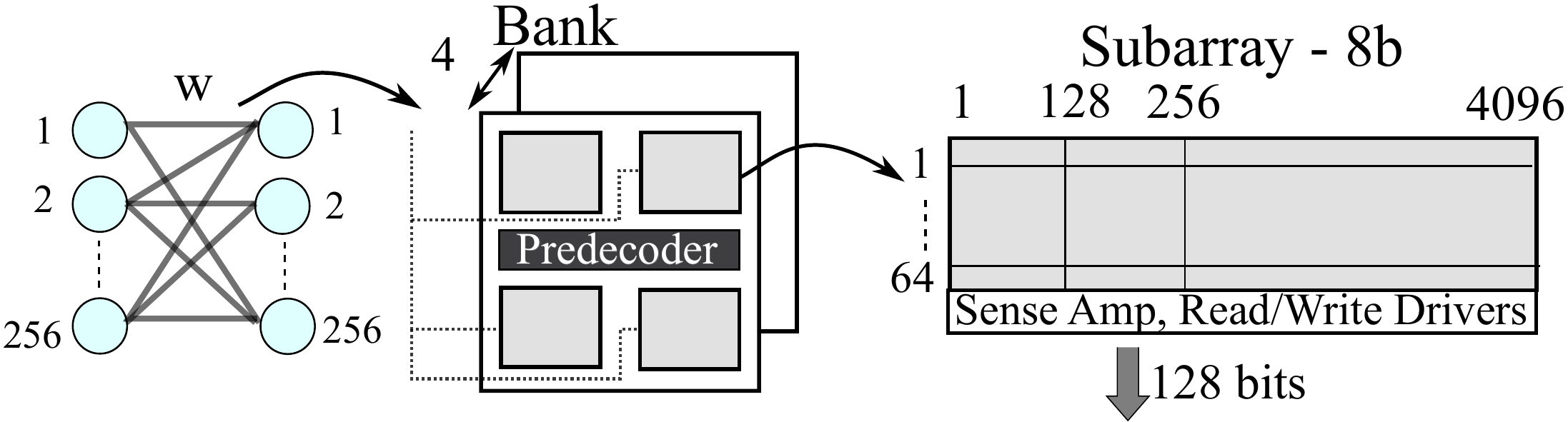}}
\caption{Banked STT-RAM organization for mapping the
synapses of GLM SNN having 256 input and 256 output neurons with 8-b precision weights.}
\label{fig:bank_mem}
\end{figure}
We describe the memory organization for the baseline design with 8-bit precision and  ($T=8$, $\tau=7$) as discussed in Section \ref{sec:alg_opt}.  The required memory capacity for any b-bit precision would become $2048\times256\,\times\,$b  bits. We design and analyze the STT-RAM based synaptic memory using DESTINY, a comprehensive tool for modeling emerging memory technologies \cite{destiny}. STT-RAM cell architecture with  1T/1MTJ configuration is simulated using the parameters mentioned in table \ref{tab:STT_RAM} with the feature size   $F=70\,$nm \cite{qualcomm_STT_RAM}. In table \ref{tab:STT_RAM}, V$_\text{r}$ is the read voltage, t$_\text{r}$ is the read pulse width, I$_{\text{p}}$ is the programming current and t$_\text{w}$ denotes the write pulse width. Representative values for read and write pulse width is assumed based on experimentally reported chip-scale demonstrations of STT-RAM \cite{STT_RAM_fastread, qualcomm_STT_RAM}. 
\begin{table}[h!]
	\caption{Simulation parameters for STT-RAM array using DESTINY with the feature size  $F=70\,$nm}
	\label{tab:STT_RAM}
	\centering 
		\begin{tabular}{|c|c|c|c|c|c|c|c|}
		\hline
			Parameter & Cell Area & R$_{\text{on}}$ & R$_{\text{off}}$ & V$_\text{r}$& t$_\text{r}$ & I$_{\text{p}}$ & t$_\text{w}$ \\
			& F$^2$ & $\Omega$ & $\Omega$ & mV & ns & $\mu$A & ns \\
			\hline
    Value & 24 & 2500 & 5000 & 80 & 5  & 150  & 10 \\
			\hline	
		\end{tabular}
\end{table}
The optimized solution of the memory mapping for 8-bit precision is shown in Figure \ref{fig:bank_mem}. The read energy per word line for $8$b is around $535\,$pJ with a latency of $7.34\,$ns. The simulated STT-RAM array running at $100\,$MHz has an area efficiency of $43$\% with a total synaptic area of $1.14\,$mm$^2$ and power of $53.5\,$mW. 
\section{Performance evaluation}
\label{sec:perf_eval}
\begin{figure}[h!]
\centerline
{\includegraphics[width=\linewidth]{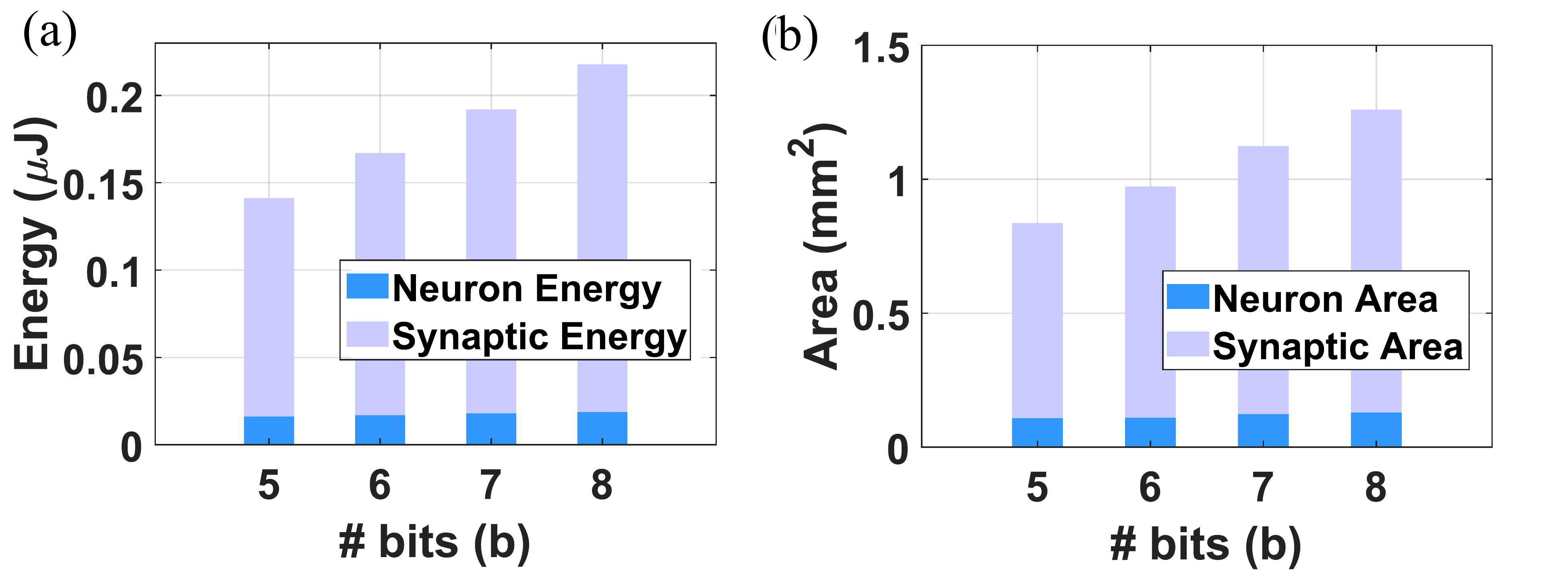}}
\caption{(a) Average energy per processor time step ($t$) of SpinAPS for each of the bit choices for the handwritten digit benchmark. One processor time step corresponds to reading all the active word lines (on an average $365$), compute the membrane potential and generating an output spike. (b) Area of the SpinAPS core as a function of bit precision.}
\label{fig:energy_mem_neuron}
\end{figure}

We use a commonly used performance metric  to benchmark our accelerator design - the number of billions of synaptic operations that can be performed per second per watt (GSOPS/W) and per unit area (GSOPS/W/mm$^{2}$). A synaptic operation refers to reading one active word line ($256$ b-bit synapses), computing the spike probability and then issuing the output spike. We compare these performance metrics of SpinAPS with an equivalent SRAM-based design for different bit precisions (shown in table \ref{table:perf_comp}), with  neuron logic implementation kept  the same. The SRAM memory is simulated using DESTINY \cite{destiny} and clocked at $250\,$MHz. Independent of bit precision, SpinAPS can perform $25.6\,$GSOPS while the SRAM-based design achieves $64\,$GSOPS due to its higher clock rate. For $8$-bit precision, SpinAPS core is approximately $6\times$ better in synaptic power and $3\times$ in synaptic area when compared to SRAM. For the values reported in table \ref{table:perf_comp}, a $10\%$ overhead is considered for the spike routing infrastructure between the cores and $20\%$ overhead is added to consider the area and power of the core's controller \cite{TrueNorth, rajit_router, DYNAPs}.   
\begin{table}[h!]
	\caption{Performance comparison of STT-RAM and SRAM-based designs.}
	\label{table:perf_comp}
	\centering 
		\begin{tabular}{|c |c|c|c|c|}
			\hline
			Precision & \multicolumn{2}{c|}{GSOPS/W} & \multicolumn{2}{c|}{GSOPS/W/mm$^2$} \\
            \cline{2-5}
             &  SRAM  &  STT-RAM & SRAM & STT-RAM\\
			\hline
			5  & 353 & 474 & 177 & 559\\
			\hline
			6   & 283 & 412 & 119 & 415\\
			\hline 
			7   & 230 & 366 & 83 & 322\\
			\hline 
			8   & 193  & 311 & 61 & 239\\
			\hline 
			
		\end{tabular}
\end{table}
It can be noticed that the SpinAPS core can achieve a maximum performance improvement of $4\times$ and a minimum of $3\times$ in terms of GSOPS/W/mm$^2$ when compared to an equivalent SRAM-based design for $8$b and $5$b precision choices respectively. The average energy per algorithmic time step $t$ of the SpinAPS core and the total area for the design as a function of the bit precision is shown in Fig. \ref{fig:energy_mem_neuron}. 

We now compare the performance of SpinAPS with an inference accelerator design employing STT-RAM devices for SNNs, memristor-based inference engines for DNNs, and GPUs. The performance improvement projected  for SpinAPS is comparable to what has been recently reported in \cite{SRK_STT_RAM}, which uses STT-RAM devices in integrated crossbar arrays. Furthermore, the SpinAPS design presented here, when extrapolated to $32\,$nm \cite{intel_bohr} technology node achieves $850\,$GSOPS/W, which is also comparable with other recent memristor-based DNN inference engines achieving $800\,$GSOPS/W in \cite{ISSAC}  and $1060\,$GSOPS/W in \cite{PUMA}. Note that when compared to state-of-the-art GPUs like Tesla V100, SpinAPS can achieve approximately $20\times$,
and $4\times$ improvement in terms of GSOPS/W, and GSOPS/mm$^2$  
 \cite{Tesla}. While implementations employing analog phase change memory (PCM) devices in crossbar arrays have been projected to provide two orders of magnitude improvement in energy efficiency when compared to GPUs through extrapolated and aggressive assumptions \cite{Ambrogio2018}, SpinAPS projections are based on more realistic  design choices that are  representative of experimental demonstrations.
\section{Conclusions}
\label{sec:conclude}
In this paper, we proposed a hardware accelerator SpinAPS using binary STT-RAM devices and digital CMOS neurons to perform inference for probabilistic SNNs. The probabilistic SNNs based on the GLM neuron model are trained using the novel first-to-spike rule and its performance is benchmarked on two standard datasets, exhibiting comparable performance with an equivalent ANN. We discussed the design of the basic elements in the SpinAPS core, considering different bit precision choices and the trade-off associated with the performance and hardware constraints. SpinAPS leverages the ability of first-to-spike rule in making decisions with low latency achieving approximately $4\times$ performance improvement in terms of GSOPS/W/mm$^2$ compared to SRAM-based design.
\section*{Acknowledgment}
This research was supported in part by the National Science Foundation grant \#1710009,  and the CAMPUSENSE project grant from CISCO Systems Inc. Resources of the High-Performance Computing facility at NJIT was used in this work.

\bibliography{References} 

\begin{thebibliography}{10}
\expandafter\ifx\csname url\endcsname\relax
  \def\url#1{\texttt{#1}}\fi
\expandafter\ifx\csname urlprefix\endcsname\relax\def\urlprefix{URL }\fi
\expandafter\ifx\csname href\endcsname\relax
  \def\href#1#2{#2} \def\path#1{#1}\fi

\bibitem{IEEE_nano}
S.~R. {Nandakumar}, et~al., Building brain-inspired computing systems:
  Examining the role of nanoscale devices, IEEE Nanotechnology Magazine 12~(3)
  (2018) 19--35.
\newblock \href {http://dx.doi.org/10.1109/MNANO.2018.2845078}
  {\path{doi:10.1109/MNANO.2018.2845078}}.

\bibitem{Roy2019}
K.~Roy, A.~Jaiswal, P.~Panda, Towards spike-based machine intelligence with
  neuromorphic computing, Nature 575~(7784) (2019) 607--617.
\newblock \href {http://dx.doi.org/10.1038/s41586-019-1677-2}
  {\path{doi:10.1038/s41586-019-1677-2}}.

\bibitem{jang_IEEE_mag}
H.~{Jang}, O.~{Simeone}, B.~{Gardner}, A.~{Gruning}, An {I}ntroduction to
  {P}robabilistic {S}piking {N}eural {N}etworks: {P}robabilistic {M}odels,
  {L}earning {R}ules, and {A}pplications, IEEE Signal Processing Magazine
  36~(6) (2019) 64--77.

\bibitem{Tijanic}
J.~Pei, et~al., Towards artificial general intelligence with hybrid tianjic
  chip architecture, Nature 572~(7767) (2019) 106--111.
\newblock \href {http://dx.doi.org/10.1038/s41586-019-1424-8}
  {\path{doi:10.1038/s41586-019-1424-8}}.

\bibitem{intel_loihi}
M.~Davies, et~al., Loihi: A {N}euromorphic {M}anycore {P}rocessor with
  {O}n-{C}hip {L}earning, IEEE Micro 38~(1) (2018) 82--99.
\newblock \href {http://dx.doi.org/10.1109/MM.2018.112130359}
  {\path{doi:10.1109/MM.2018.112130359}}.

\bibitem{TrueNorth}
P.~A. Merolla, J.~V. Arthur, et~al., A million spiking-neuron integrated
  circuit with a scalable communication network and interface, Science
  345~(6197) (2014) 668--673.
\newblock \href {http://dx.doi.org/10.1126/science.1254642}
  {\path{doi:10.1126/science.1254642}}.

\bibitem{Kuzum_2013}
D.~Kuzum, S.~Yu, H.-S.~P. Wong, Synaptic electronics: materials, devices and
  applications, Nanotechnology 24~(38).
\newblock \href {http://dx.doi.org/10.1088/0957-4484/24/38/382001}
  {\path{doi:10.1088/0957-4484/24/38/382001}}.

\bibitem{RPU}
T.~Gokmen, Y.~Vlasov, Acceleration of {D}eep {N}eural {N}etwork {T}raining with
  {R}esistive {C}ross-{P}oint {D}evices: {D}esign {C}onsiderations, Frontiers
  in Neuroscience 10 (2016) 333.
\newblock \href {http://dx.doi.org/10.3389/fnins.2016.00333}
  {\path{doi:10.3389/fnins.2016.00333}}.

\bibitem{Ambrogio2018}
S.~Ambrogio, et~al., Equivalent-accuracy accelerated neural-network training
  using analogue memory, Nature 558.
\newblock \href {http://dx.doi.org/10.1038/s41586-018-0180-5}
  {\path{doi:10.1038/s41586-018-0180-5}}.

\bibitem{PUMA}
A.~Ankit, I.~E. Hajj, et~al., {PUMA}: {A} {P}rogrammable {U}ltra-{E}fficient
  {M}emristor-{B}ased {A}ccelerator for {M}achine {L}earning {I}nference, in:
  Proceedings of the Twenty-Fourth International Conference on Architectural
  Support for Programming Languages and Operating Systems, ASPLOS ’19,
  Association for Computing Machinery, New York, NY, USA, 2019, p. 715–731.
\newblock \href {http://dx.doi.org/10.1145/3297858.3304049}
  {\path{doi:10.1145/3297858.3304049}}.

\bibitem{ISSAC}
A.~{Shafiee}, A.~{Nag}, N.~{Muralimanohar}, R.~{Balasubramonian}, J.~P.
  {Strachan}, M.~{Hu}, R.~S. {Williams}, V.~{Srikumar}, {ISAAC}: {A}
  {C}onvolutional {N}eural {N}etwork {A}ccelerator with {I}n-{S}itu {A}nalog
  {A}rithmetic in {C}rossbars, in: 2016 ACM/IEEE 43rd Annual International
  Symposium on Computer Architecture (ISCA), 2016, pp. 14--26.

\bibitem{PRIME}
P.~{Chi}, S.~{Li}, C.~{Xu}, T.~{Zhang}, J.~{Zhao}, Y.~{Liu}, Y.~{Wang},
  Y.~{Xie}, {PRIME}: {A} {N}ovel {P}rocessing-in-{M}emory {A}rchitecture for
  {N}eural {N}etwork {C}omputation in {ReRAM-Based Main Memory}, in: 2016
  ACM/IEEE 43rd Annual International Symposium on Computer Architecture (ISCA),
  2016, pp. 27--39.

\bibitem{Anakha_neruo_comp}
A.~V. Babu, S.~Lashkare, U.~Ganguly, B.~Rajendran, {Stochastic Learning in Deep
  Neural Networks based on Nanoscale PCMO device characteristics},
  Neurocomputing 321 (2018) 227 -- 236.
\newblock \href
  {http://dx.doi.org/https://doi.org/10.1016/j.neucom.2018.09.019}
  {\path{doi:https://doi.org/10.1016/j.neucom.2018.09.019}}.

\bibitem{SRK_STT_RAM}
S.~R. {Kulkarni}, D.~V. {Kadetotad}, S.~{Yin}, J.~{Seo}, B.~{Rajendran},
  Neuromorphic hardware accelerator for snn inference based on stt-ram crossbar
  arrays, in: 2019 26th IEEE International Conference on Electronics, Circuits
  and Systems (ICECS), 2019, pp. 438--441.

\bibitem{STTRAM_synapse}
N.~{Locatelli}, et~al., Spintronic devices as key elements for energy-efficient
  neuroinspired architectures, in: 2015 Design, Automation Test in Europe
  Conference Exhibition (DATE), 2015, pp. 994--999.
\newblock \href {http://dx.doi.org/10.7873/DATE.2015.1117}
  {\path{doi:10.7873/DATE.2015.1117}}.

\bibitem{STT_RAM_synapse1}
Y.~{Wang}, et~al., A case of precision-tunable {STT-RAM} memory design for
  approximate neural network, in: 2015 IEEE International Symposium on Circuits
  and Systems (ISCAS), 2015.
\newblock \href {http://dx.doi.org/10.1109/ISCAS.2015.7168938}
  {\path{doi:10.1109/ISCAS.2015.7168938}}.

\bibitem{Sengupta_IJCNN}
A.~{Sengupta}, A.~{Ankit}, K.~{Roy}, Performance analysis and benchmarking of
  all-spin spiking neural networks (special session paper), in: 2017
  International Joint Conference on Neural Networks (IJCNN), 2017, pp.
  4557--4563.
\newblock \href {http://dx.doi.org/10.1109/IJCNN.2017.7966434}
  {\path{doi:10.1109/IJCNN.2017.7966434}}.

\bibitem{SRK_STT_RAM_learn}
S.~R. {Kulkarni}, S.~{Yin}, J.~{Seo}, B.~{Rajendran}, {An On-Chip Learning
  Accelerator for Spiking Neural Networks using STT-RAM Crossbar Arrays}, in:
  2020 Design, Automation Test in Europe Conference Exhibition (DATE), 2020,
  pp. 1019--1024.

\bibitem{STT_RAM_ANNs}
S.~Fukami, H.~Ohno, Perspective: Spintronic synapse for artificial neural
  network, Journal of Applied Physics 124~(15) (2018) 151904.
\newblock \href {http://dx.doi.org/10.1063/1.5042317}
  {\path{doi:10.1063/1.5042317}}.

\bibitem{STTRAM_prob_SNN}
A.~{Sengupta}, M.~{Parsa}, B.~{Han}, K.~{Roy}, Probabilistic {D}eep {S}piking
  {N}eural {S}ystems {E}nabled by {M}agnetic {T}unnel {J}unction, IEEE
  Transactions on Electron Devices 63~(7) (2016) 2963--2970.
\newblock \href {http://dx.doi.org/10.1109/TED.2016.2568762}
  {\path{doi:10.1109/TED.2016.2568762}}.

\bibitem{SNN_realize1}
B.~{Rueckauer}, S.~{Liu}, Conversion of analog to spiking neural networks using
  sparse temporal coding, in: 2018 IEEE International Symposium on Circuits and
  Systems (ISCAS), 2018, pp. 1--5.

\bibitem{SNN_realize2}
Y.~Wu, L.~Deng, G.~Li, J.~Zhu, L.~Shi, {Spatio-Temporal Backpropagation for
  Training High-Performance Spiking Neural Networks}, Frontiers in Neuroscience
  12 (2018) 331.
\newblock \href {http://dx.doi.org/10.3389/fnins.2018.00331}
  {\path{doi:10.3389/fnins.2018.00331}}.

\bibitem{SNN_realize3}
J.~H. Lee, T.~Delbruck, M.~Pfeiffer, Training deep spiking neural networks
  using backpropagation, Frontiers in Neuroscience 10 (2016) 508.
\newblock \href {http://dx.doi.org/10.3389/fnins.2016.00508}
  {\path{doi:10.3389/fnins.2016.00508}}.

\bibitem{SNN_realize4}
P.~O'Connor, M.~Welling, {Deep Spiking Networks}, CoRR abs/1602.08323.
\newblock \href {http://arxiv.org/abs/1602.08323} {\path{arXiv:1602.08323}}.

\bibitem{Pillow_GLM}
A.~I. Weber, J.~W. Pillow, Capturing the {D}ynamical {R}epertoire of {S}ingle
  {N}eurons with {G}eneralized {L}inear {M}odels, Neural Computation\href
  {http://dx.doi.org/10.1162/neco_a_01021} {\path{doi:10.1162/neco_a_01021}}.

\bibitem{Alireza_FTS}
A.~{Bagheri}, et~al., Training {P}robabilistic {S}piking {N}eural {N}etworks
  with {F}irst- {T}o-{S}pike {D}ecoding, in: 2018 IEEE International Conference
  on Acoustics, Speech and Signal Processing (ICASSP), 2018.

\bibitem{Jang_Osvaldo}
H.~{Jang}, O.~{Simeone}, {Training {D}ynamic {E}xponential {F}amily {M}odels
  with {C}ausal and {L}ateral {D}ependencies for {G}eneralized {N}euromorphic
  {C}omputing}, in: ICASSP 2019 - 2019 IEEE International Conference on
  Acoustics, Speech and Signal Processing (ICASSP), 2019.
\newblock \href {http://dx.doi.org/10.1109/ICASSP.2019.8682960}
  {\path{doi:10.1109/ICASSP.2019.8682960}}.

\bibitem{gardner2016supervised}
B.~Gardner, A.~Grüning, Supervised learning in spiking neural networks for
  precise temporal encoding, PLOS ONE 11~(8) (2016) 1--28.
\newblock \href {http://dx.doi.org/10.1371/journal.pone.0161335}
  {\path{doi:10.1371/journal.pone.0161335}}.

\bibitem{HAR}
D.~Anguita, A.~Ghio, L.~Oneto, X.~Parra, J.~L. Reyes-Ortiz, {Energy Efficient
  Smartphone-Based Activity Recognition using Fixed-Point Arithmetic.}, J. UCS
  19~(9) (2013) 1295--1314.

\bibitem{Sigmoid_bkgnd}
M.~T. {Tommiska}, {Efficient Digital Implementation of the Sigmoid function for
  Reprogrammable Logic}, IEE Proceedings - Computers and Digital Techniques
  150~(6) (2003) 403--411.
\newblock \href {http://dx.doi.org/10.1049/ip-cdt:20030965}
  {\path{doi:10.1049/ip-cdt:20030965}}.

\bibitem{Sigmoid_PWL}
C.~Alippi, G.~Storti-Gajani, Simple approximation of sigmoidal functions:
  realistic design of digital neural networks capable of learning, in: IEEE
  International Sympoisum on Circuits and Systems, 1991.
\newblock \href {http://dx.doi.org/10.1109/ISCAS.1991.176661}
  {\path{doi:10.1109/ISCAS.1991.176661}}.

\bibitem{STT-RAM_design}
R.~{Dorrance}, et~al., {S}calability and {D}esign-{S}pace {A}nalysis of a
  {1T-1MTJ} {M}emory {C}ell for {STT-RAM}s, IEEE Transactions on Electron
  Devices 59~(4) (2012) 878--887.
\newblock \href {http://dx.doi.org/10.1109/TED.2011.2182053}
  {\path{doi:10.1109/TED.2011.2182053}}.

\bibitem{STT_RAM_40nm}
H.~{Yu}, et~al., {C}ycling endurance optimization scheme for {1Mb STT-MRAM} in
  40nm technology, in: 2013 IEEE International Solid-State Circuits Conference
  Digest of Technical Papers, 2013, pp. 224--225.
\newblock \href {http://dx.doi.org/10.1109/ISSCC.2013.6487710}
  {\path{doi:10.1109/ISSCC.2013.6487710}}.

\bibitem{STT_RAM_90nm}
R.~{Nebashi}, et~al., A 90nm 12ns {32Mb} {2T1MTJ MRAM}, in: 2009 IEEE
  International Solid-State Circuits Conference - Digest of Technical Papers,
  2009, pp. 462--463,463a.
\newblock \href {http://dx.doi.org/10.1109/ISSCC.2009.4977508}
  {\path{doi:10.1109/ISSCC.2009.4977508}}.

\bibitem{STT_RAM_fastread}
Q.~{Dong}, et~al., A 1-{M}b 28-nm {1T1MTJ STT-MRAM} with {S}ingle-{C}ap
  {O}ffset-{C}ancelled {S}ense {A}mplifier and {I}n {S}itu
  {S}elf-{W}rite-{T}ermination, IEEE Journal of Solid-State Circuits 54~(1)
  (2019) 231--239.
\newblock \href {http://dx.doi.org/10.1109/JSSC.2018.2872584}
  {\path{doi:10.1109/JSSC.2018.2872584}}.

\bibitem{STT_RAM_3_3ns}
H.~{Noguchi}, et~al., A 3.3ns-access-time {71.2$\,\mu$W/MHz} {1Mb} embedded
  {STT-MRAM} using physically eliminated read-disturb scheme and normally-off
  memory architecture, in: IEEE International Solid-State Circuits Conference -
  (ISSCC) Digest of Technical Papers, 2015.
\newblock \href {http://dx.doi.org/10.1109/ISSCC.2015.7062963}
  {\path{doi:10.1109/ISSCC.2015.7062963}}.

\bibitem{destiny}
M.~Poremba, et~al., {DESTINY}: A tool for modeling emerging 3{D NVM and eDRAM}
  caches, in: 2015 Design, Automation Test in Europe Conference Exhibition
  (DATE), 2015.
\newblock \href {http://dx.doi.org/10.7873/DATE.2015.0733}
  {\path{doi:10.7873/DATE.2015.0733}}.

\bibitem{qualcomm_STT_RAM}
C.~J. {Lin}, et~al., 45nm low power {CMOS} logic compatible embedded {STT MRAM}
  utilizing a reverse-connection {1T/1MTJ} cell, in: 2009 IEEE International
  Electron Devices Meeting (IEDM), 2009, pp. 1--4.

\bibitem{rajit_router}
S.~Moradi, R.~Manohar, The {I}mpact of {O}n-chip {C}ommunication on {M}emory
  {T}echnologies for {N}euromorphic {S}ystems, Journal of Physics D: Applied
  Physics 52~(1) (2018) 014003.
\newblock \href {http://dx.doi.org/10.1088/1361-6463/aae641}
  {\path{doi:10.1088/1361-6463/aae641}}.

\bibitem{DYNAPs}
S.~{Moradi}, et~al., A {S}calable {M}ulticore {A}rchitecture with
  {H}eterogeneous {M}emory {S}tructures for {D}ynamic {N}euromorphic
  {A}synchronous {P}rocessors ({DYNAPs}), IEEE Transactions on Biomedical
  Circuits and Systems\href {http://dx.doi.org/10.1109/TBCAS.2017.2759700}
  {\path{doi:10.1109/TBCAS.2017.2759700}}.

\bibitem{intel_bohr}
M.~Bohr, 14 nm {P}rocess {T}echnology: {O}pening {N}ew {H}orizons, intel
  {T}echnology {D}evelopment, Available online.
  \url{https://www.intel.com/content/www/us/en/architecture-and-technology/bohr-14nm-idf-2014-brief.html}
  (2014).

\bibitem{Tesla}
{NVIDIA} {V}olta {GV}100 12nm {FinFET} {GPU} {D}etailed-{T}esla {V}100
  {S}pecifications {I}nclude 21 {B}illion {T}ransistors, 5120{CUDA} {C}ores,
  16{GB} {HBM2} with 900 {GB/}s {B}andwidth, available online. \url
  {https://wccftech.com/nvidia-volta-gv100-gpu-tesla-v100-architecture-
  specifications-deep-dive/}.

\end{thebibliography}
\ignore{  
\section*{Vitae} 
\begin{wrapfigure}{L}{2cm}
\end{wrapfigure} 
Anakha V Babu  received the B.Tech degree in electronics and communication engineering from University of Kerala, India, in 2010 and M.tech degree in Microelectronics \& VLSI design from the Indian Institute of Technology Bombay, Mumbai, India, in 2016. She is currently pursuing Ph.D degree in electrical engineering at the New Jersey Institute of Technology, New Jersey, USA. Her current research interests include bio-inspired computing using novel memristive devices.
 }
\ignore{
\begin{wrapfigure}{L}{2cm}
\includegraphics[width=2cm]{./Images/bipin_v1}
\end{wrapfigure}
Bipin Rajendran received a B.\,Tech degree from I.I.T.~Kharagpur   in  2000,  and    M.S.~and  Ph.D.~degrees in Electrical Engineering from Stanford University   in  2003  and  2006,  respectively. He was a Master Inventor and Research Staff Member at IBM T. J. Watson Research Center in New York during 2006-'12 and a faculty member in the  Electrical  Engineering  Department at I.I.T.  Bombay during 2012-'15.   His research focuses   on   building   algorithms, devices   and   systems   for brain-inspired  computing.  He  has  authored  over  60 papers  in  peer-reviewed  journals  and  conferences, and has been issued  55 U.S. patents. He is currently an Associate Professor   of Electrical \& Computer Engineering  at New Jersey Institute of Technology. 
}

\end{document}